\documentclass[sigconf,nonacm]{acmart}
\usepackage{pvldb}
\usepackage{algorithm,algorithmic,array}
\usepackage{xurl}
\renewcommand\vldbdoi{XX.XX/XXX.XX}
\renewcommand\vldbpages{XXX-XXX}
\renewcommand\vldbavailabilityurl{https://github.com/SSMa913/NICLab-TRESS}

\begin{document}
\title{TRESS: A Fixed-Size History Representation for Irregular Time Series Forecasting}
\author{Shusen Ma}
\affiliation{\department{Institute of Advanced Technology}
  \institution{University of Science and Technology of China}
  \city{Hefei}\country{China}}

\author{Yangchun Xie}
\affiliation{\department{Institute of Advanced Technology}
  \institution{University of Science and Technology of China}
  \city{Hefei}\country{China}}

\author{Xudong Zuo}
\affiliation{\department{Institute of Advanced Technology}
  \institution{University of Science and Technology of China}
  \city{Hefei}\country{China}}

\author{Haizhou Liang}
\affiliation{\department{Institute of Advanced Technology}
  \institution{University of Science and Technology of China}
  \city{Hefei}\country{China}}
  
\author{Yun-Bo Zhao}
\authornote{Corresponding author.}
\affiliation{\department{Department of Automation}
  \institution{University of Science and Technology of China}
  \city{Hefei}\country{China}}
\email{ybzhao@ustc.edu.cn}
\begin{abstract}
A fixed number of recent observations may span minutes for one variable and days for another. Earlier measurements can still reveal whether the latest value is unusual for that variable. We propose TRESS (Timestamped Recent Events and Statistical Summaries), a forecasting model that retains the latest observations at their original times and summarises the full history of each variable. For a given set of variables and event budget, this representation stays the same size as the record grows. Its query-conditioned event lifting module weights deviations from the latest value according to the age of each observation and the forecast horizon. A channel-specific decoder combines the resulting features with historical statistics and a summary shared across variables to predict a correction to the latest value. Retaining 8 to 16 observations per variable, TRESS records the lowest mean test MSE and MAE among the methods compared on all four benchmarks. Serving 32 successive query times from one encoded history reduces CPU inference time by a factor of 2.12--2.60 across the four datasets. The code is at https://github.com/SSMa913/NICLab-TRESS.
\end{abstract}
\maketitle


\section{Introduction}
\label{sec:introduction}

\begin{figure*}[!t]
\centering
\includegraphics[width=0.85\linewidth]{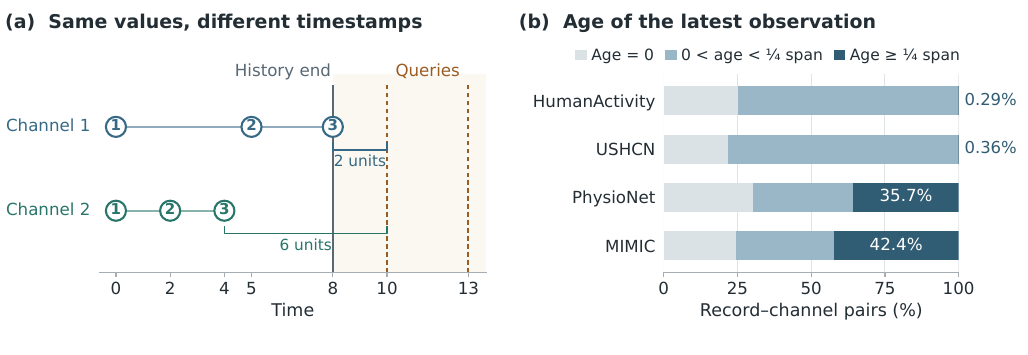}
\caption{Observation timing in irregular histories. In the constructed example, numbers inside circles give the observed values. The brackets measure elapsed time from each channel's last observation to the first query. The right panel summarizes the training histories. Each channel observed at least once in a record contributes one age, measured from its latest observation to the common history endpoint. Age is expressed relative to the record's historical span $S$, as defined in Section~\ref{sec:formulation}.}
\Description{Two example channels share the same values at different times. A bar chart reports differences in the age of the latest observation across four training sets.}
\label{fig:motivation}
\end{figure*}

Forecasting from irregular observations requires knowing both what was measured and when. Clinical variables are often recorded at different times, and gaps also arise in activity and environmental measurements~\cite{grud,yalavarthi2024grafiti}. Consider the two channels in Figure~\ref{fig:motivation}. Both take the values $1,2,3$ in order. At time $10$, their latest measurements are two and six time units old, respectively. Moving the query from $10$ to $13$ increases the common forecast horizon from two to five units while the ages of the observations at the history endpoint stay fixed. When forecasting channel 2, a joint model can also use the two observations of channel 1 that arrived after channel 2 was last measured.

We measure the age of each channel's latest observation at the common history endpoint in the four training sets. In PhysioNet and MIMIC, respectively, 35.7\% and 42.4\% of observed record--channel pairs have a latest measurement at least a quarter of the record's history span before that endpoint. The corresponding fractions on HumanActivity and USHCN are both below 0.4\%.

\begin{figure}[!t]
\centering
\includegraphics[width=\columnwidth]{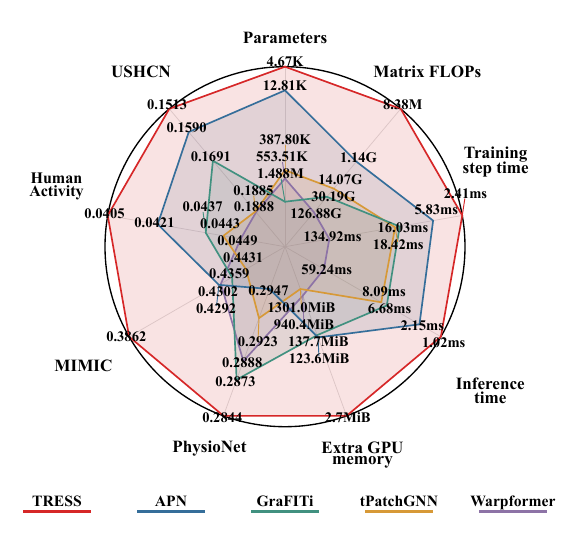}
\caption{MSE on each dataset (Table~\ref{tab:accuracy}) and computational costs averaged arithmetically over four datasets. Numbers show actual values; K/M/G denote $10^3/10^6/10^9$. Times are milliseconds per batch ($b=32$); training includes forward, backward, and Adam. Memory is extra peak inference allocation in MiB. Matrix FLOPs count profiled matrix products and convolutions. Each axis is independently scaled in log space, with lower values farther out. Figure~\ref{fig:efficiency} reports the costs for each dataset.}
\Description{Radar chart comparing five models on test MSE for each dataset and on five costs averaged across datasets.}
\label{fig:accuracy-efficiency}
\end{figure}

A forecast over an event table fixes an observation cutoff and asks for values at one or more later times. When a clinician requests another forecast for the same patient record, the requested time can change while the available history stays exactly the same. This raises a practical question about where to do the work. The model needs the times and values of recent measurements for each variable, but repeated requests need not reread a long history to recover them. We examine which inputs can be retained once per record and measure the cost of serving additional query times from that representation.

Prior work on irregular observations uses continuous-time latent states~\cite{schirmer2022cru,neuralflows}, timestamp-based attention~\cite{mtan}, and graph-based models~\cite{yalavarthi2024grafiti,tpatchgnn}. We study whether making event timing explicit permits accurate forecasts with a small shared network.

Small models are competitive on regularly sampled forecasting benchmarks. DLinear and TSMixer use linear or MLP predictors~\cite{zeng2023linear,chen2023tsmixer}; SparseTSF and CycleNet further exploit periodic structure~\cite{lin2024sparsetsf,lin2024cyclenet}. Their index-based inputs do not distinguish short and long gaps between measurements. APN addresses irregular forecasting with time-aware adaptive patch aggregation and a shallow prediction network~\cite{liu2026rethinking}. TRESS retains individual recent events and conditions their contributions on the prediction query.

Observation age and forecast distance have distinct roles. A deviation that describes a recent change may become less useful farther into the future. Variables also differ in how much they depend on their own histories and on observations from other variables. We therefore separate the construction of local temporal features from the channel-specific use of those features and shared context.

Figure~\ref{fig:tress-framework} presents \textbf{TRESS} (Timestamped Recent Events and Statistical Summaries). Query-conditioned event lifting (QEL) centers recent measurements on each variable's latest value. An event gate weights these deviations using historical age and forecast distance. A history projection combines the gated events with event lags, validity indicators, and history statistics; a query projection adds information about the requested future time. The statistics preserve level, spread, and sampling density beyond the retained events.

Channel-conditioned joint decoding (CJD) uses channel modulation and a shared residual block to form the local representation. In parallel, a context projection summarizes the joint history, and an age-count gate controls its contribution to each channel and query. The local representation, four direct statistics, and gated context are concatenated into joint features for one channel readout. Its latest-value coefficient is learned around one, so training starts from persistence. For fixed feature widths and event budget, the number of learned parameters is independent of history length and the number of forecast queries.

TRESS achieves the lowest mean test MSE and MAE on HumanActivity, USHCN, PhysioNet, and MIMIC among the methods in Table~\ref{tab:accuracy}. Figure~\ref{fig:accuracy-efficiency} compares dataset-level MSE and mean computational costs for TRESS and four baselines. TRESS uses fewer parameters and less additional GPU memory on every dataset, while runtime gains depend on the workload.

Our contributions are as follows:
\begin{itemize}
\item We propose \textbf{TRESS}, a compact model for irregular time series forecasting at specified future query times. With 1,574--9,457 parameters, TRESS has the lowest reported mean MSE and MAE on all four benchmarks in Table~\ref{tab:accuracy}.
\item We introduce \textbf{query-conditioned event lifting}, which weights recent event deviations using historical age and forecast distance, then combines history and query projections. The history input retains event order, availability, and summary statistics.
\item We introduce \textbf{channel-conditioned joint decoding} to combine local features, direct statistics, and gated context in a single readout. Channel modulation, a shared residual block, and an age-count gate adapt these features with few additional parameters; the latest-value coefficient is learned around one.
\item We characterize the history that suffices for forecast queries on a record and evaluate how much computation can be shared when the client requests additional forecast times. We also measure the contribution of cross-variable context separately for recent, older, and previously unobserved target variables.

\end{itemize}

\section{Related Work}
\label{sec:related-work}

\paragraph{Irregular observations and continuous time.}
Irregular records require models to account for unequal gaps within a variable and misaligned observations across variables. Continuous-time approaches represent temporal evolution between measurements. Log-NCDE uses the Log-ODE method and Lie-bracket structure to approximate neural controlled differential equations~\cite{walker2024logncde}. BayOTIDE places Gaussian-process priors on low-rank temporal factors and converts them to a state-space representation for online imputation at arbitrary timestamps~\cite{fang2024bayotide}. Recent point predictors also use explicit aggregation. APN learns temporal patch boundaries and time-aware weighted summaries before query aggregation and a shallow predictor~\cite{liu2026rethinking}. KAFNet retains canonical pre-alignment and combines pre-convolution, temporal kernel aggregation, and frequency-domain linear attention~\cite{zhou2026kafnet}. TRESS builds query-conditioned features directly from recent observations and compact history summaries.

\paragraph{Interactions across observations and queries.}
GraFITi represents each observation as an edge joining a variable and a timestamp~\cite{yalavarthi2024grafiti}, while tPatchGNN learns time-adaptive graphs over patch representations~\cite{tpatchgnn}. HyperIMTS treats observations as nodes connected by temporal and variable hyperedges~\cite{li2025hyperimts}. TimeCHEAT models channels jointly within a patch and applies attention across patches independently for each channel~\cite{liu2025timecheat}. ReIMTS recursively partitions histories into shorter intervals while preserving timestamps, then fuses representations across scales~\cite{li2026reimts}. Probabilistic methods additionally model dependence among forecast targets: ProFITi learns conditional joint distributions over queried channels and future times from irregular history using normalizing flows~\cite{yalavarthi2025profiti}, and MOSES uses mixtures of Gaussian-based separable flows to make marginal and joint forecasts consistent~\cite{yalavarthi2026moses}. TRESS uses a shared context projection for cross-variable information; channel modulation and an age-count gate adapt the local representation and shared context before a channel-specific point readout.

\paragraph{Compact forecasting architectures.}
Linear and MLP forecasters provide useful precedents for compact prediction. DLinear separates trend and residual components, while NLinear centers each history on its latest value~\cite{zeng2023linear}. SparseTSF exploits cross-period subsampling, and CycleNet predicts deviations from learned recurring patterns~\cite{lin2024sparsetsf,lin2024cyclenet}. MoLE routes inputs among linear experts~\cite{ni2024mole}; TSMixer alternates temporal and feature mixing~\cite{chen2023tsmixer}. TRESS shares NLinear's latest-value reference, while retaining variable-specific observation times and learning the latest-value readout coefficient around one.

\paragraph{Temporal structure and shared representations.}
TimeMixer mixes seasonal and trend components across scales~\cite{wang2024timemixer}. ModernTCN uses large-receptive-field convolutions~\cite{luo2024moderntcn}. FITS interpolates in the complex frequency domain, while TimeKAN models frequency bands with Kolmogorov--Arnold network (KAN) blocks~\cite{xu2024fits,huang2025timekan}; FreDF introduces a frequency-domain training objective~\cite{wang2025fredf}. These methods exploit temporal structure in regularly sampled inputs. TRESS uses observed timestamps to form history and query coordinates directly.

For cross-variable communication, SOFTS aggregates and redistributes a shared core~\cite{han2024softs}, iTransformer attends over variable tokens~\cite{liu2024itransformer}, and TimeXer links temporal patches to exogenous-variable representations~\cite{wang2024timexer}. TRESS projects history summaries into a shared context and gates it by observation age and count. Pretraining learns representations transferable across datasets. MOMENT reconstructs masked patches~\cite{goswami2024moment}, Timer and TimesFM use generative forecasting objectives~\cite{liu2024timer,das2024timesfm}, and Timer-XL extends next-token prediction to multivariate contexts~\cite{liu2025timerxl}. Chronos encodes quantized observations as tokens~\cite{ansari2024chronos}, while Moirai employs patch projections and attention across time and variables~\cite{woo2024moirai}. TRESS fits one compact predictor per dataset.

\paragraph{Historical data and repeated queries.}
ModelarDB stores segments of a time series as models and evaluates historical aggregates over the segments~\cite{jensen2018modelardb}. The representation must therefore support questions about past values. TRESS has a narrower target: given an observation window, recover the inputs needed to predict at a future time. The recent events and grouped statistics in Section~\ref{sec:state} suffice for that prediction, even though they omit the individual values of older measurements. The distinction matters when deciding which downstream requests a stored representation can answer.

DBToaster materializes intermediate views and updates them as the data change~\cite{ahmad2012dbtoaster}. In our setting, two kinds of change have different costs. A new measurement can change the stored historical state. A new forecast time for the same snapshot changes only the query-dependent calculation. 
Section~\ref{sec:query-scaling} holds the origin fixed and varies the number and scheduling of future times. This measures how much work can be shared across requests.

\section{Method}
\label{sec:method}

\begin{figure*}[t]
\centering
\includegraphics[width=\textwidth]{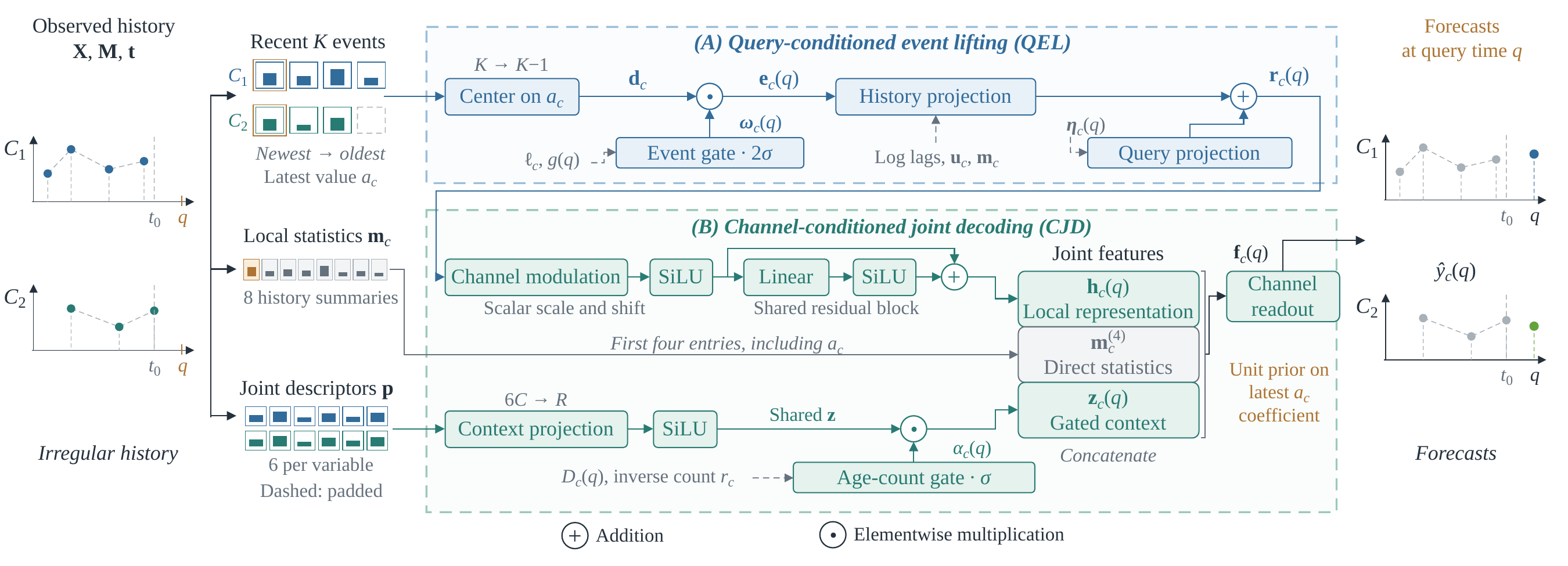}
\caption{TRESS forecasts variable $c$ at query time $q$ from historical values $\mathbf X$, mask $\mathbf M$, and timestamps $\mathbf t$. QEL multiplies deviations $\mathbf d_c$ from the latest value $a_c$ by event weights $\boldsymbol\omega_c(q)$, both over slots $2,\ldots,K$, then combines them with log lags, support indicators $\mathbf u_c$, and eight statistics $\mathbf m_c$. History and query projections are added to obtain $\mathbf r_c(q)$; query features $\boldsymbol\eta_c(q)$ derive from $q$, the history endpoint $t_0$, span $S$, and event lags. CJD forms $\mathbf h_c(q)$ through channel modulation and a residual block. Its scalar gate $\alpha_c(q)$ modulates context $\mathbf z$ projected from joint descriptors $\mathbf p$. The readout combines $\mathbf h_c(q)$, four direct statistics $\mathbf m_c^{(4)}$, and gated context $\mathbf z_c(q)$, with a unit prior on the latest-value weight. Dashed module outlines group computations; dashed event slots denote padding.}
\Description{Two connected modules encode recent observations relative to each forecast time and decode them with statistics and gated cross-variable information.}
\label{fig:tress-framework}
\end{figure*}

TRESS comprises two modules (Figure~\ref{fig:tress-framework}). Query-conditioned event lifting (QEL) centers recent events, gates their deviations, and adds history and query projections. Channel-conditioned joint decoding (CJD) combines a local representation, direct statistics, and gated context in a channel readout. Algorithm~\ref{alg:tress} gives the forward pass.

\subsection{Problem formulation}
\label{sec:formulation}

Let $\mathbf{X}\in\mathbb{R}^{L\times C}$ contain $C$ variables at $L$ historical timestamps $\mathbf{t}=(t_1,\ldots,t_L)$. The binary mask $\mathbf{M}\in\{0,1\}^{L\times C}$ has $M_{ic}=1$ when variable $c$ is observed at $t_i$; masked values are ignored. Given $Q$ future times $\mathbf{q}=(q_1,\ldots,q_Q)$, we predict $\widehat{\mathbf{Y}}\in\mathbb{R}^{Q\times C}$. Timestamps may be irregular and observations asynchronous. We describe one record and query $q$; records and queries are processed in parallel.

The latest and earliest timestamps with any observation are $t_0$ and $t_{\min}$; the historical span is $S=\max(t_0-t_{\min},0.05)$. The \emph{Recent $K$ events} input gathers each variable's observations in reverse chronological order and pads missing slots. Slot $j$ has value $x_{cj}$, time $\tau_{cj}$, and validity $u_{cj}\in\{0,1\}$. The latest value is $a_c=x_{c1}$ at $t_c^*=\tau_{c1}$. The normalized event lag and future gap are
\begin{equation}
\ell_{cj}=\frac{t_0-\tau_{cj}}{S},\qquad
g(q)=\max\!\left(\frac{q-t_0}{S},0\right).
\label{eq:time-coordinates}
\end{equation}
We set padded lags to zero. The support vector $\mathbf{u}_c$ distinguishes these slots from observed events at $t_0$. An entirely unobserved variable receives zero-valued event slots, lags and historical moments. Its last timestamp defaults to $t_0$, and it can still draw on observations from other variables.

\subsection{Query-conditioned event lifting}
\label{sec:qel}

\paragraph{Centering.}
The \emph{Center on $a_c$} block expresses observed events relative to the latest value:
\begin{equation}
d_{cj}=u_{cj}(x_{cj}-a_c).
\end{equation}
Since $d_{c1}=0$, centering yields the $K-1$ nontrivial deviations $\mathbf d_c=[d_{c2},\ldots,d_{cK}]^\top$.

\paragraph{Event gate.}
Historical age and forecast distance receive separate coefficients. Define $p_{cj}=\log(1+s_t\ell_{cj})$ and $f(q)=\log(1+s_tg(q))$, with fixed scale $s_t=16$. The gate and its multiplication with each deviation are
\begin{equation}
\begin{aligned}
\omega_{cj}(q)&=2\sigma\!\left(\sqrt{2}\theta_{j0}
 +(\theta_{j1}-\kappa)p_{cj}\right.\\[-2pt]
 &\hspace{44pt}\left. +(\theta_{j2}+\kappa)f(q)\right),\\
e_{cj}(q)&=\omega_{cj}(q)d_{cj},
\end{aligned}
\label{eq:event-gate}
\end{equation}
where $\sigma$ is sigmoid, $\theta_{jr}$ ($r\in\{0,1,2\}$) are learned per event position and shared across variables, and $\kappa\geq0$ is a fixed offset. For $\kappa>0$, zero initialization favors recent events; both age and horizon effects remain learnable. The factor two permits attenuation or amplification. In Figure~\ref{fig:tress-framework}, multiplication applies one gate per deviation. The resulting vector $\mathbf{e}_c(q)$ has $K-1$ entries.

\paragraph{History projection.}
Gated events enter the history projection together with log lags, support flags, and the \emph{Local statistics} input:
\begin{equation}
\begin{aligned}
\mathbf{m}_c=[&a_c,\ \mu_c-a_c,\ \bar{x}_c-a_c,\ v_c-a_c,\\
              &\sigma_c,\ \pi_c,\ r_c,\ \ell_{c1}]^\top.
\end{aligned}
\label{eq:metadata}
\end{equation}
Here $\mu_c$, $v_c$, and $\sigma_c$ are the full-history mean, minimum, and population standard deviation; $\bar{x}_c$ is the retained-event mean; $\pi_c$ is the fraction at the historical minimum; and $r_c=1/(1+n_c)$ for observation count $n_c$. Only observed entries contribute. These summaries retain information beyond the recent events.

\paragraph{Query projection and addition.}
The query projection uses the future gap and elapsed time since channel $c$ was last observed. The two projections meet at the addition node:
\begin{equation}
\begin{aligned}
\mathbf{v}_c(q)&=[\mathbf{e}_c(q);\log(1+16\boldsymbol{\ell}_c);
                         \mathbf{u}_c;\mathbf{m}_c],\\
D_c(q)&=\log\!\left(1+16[g(q)+\ell_{c1}]\right),\\
\boldsymbol{\eta}_c(q)&=[g(q),D_c(q),D_c(q)^2]^\top,\\
\mathbf{r}_c(q)&=\mathbf{W}_v\mathbf{v}_c(q)
                    +\mathbf{W}_q\boldsymbol{\eta}_c(q)+\mathbf{b}_v.
\end{aligned}
\label{eq:shared-projection}
\end{equation}
The brackets concatenate features. The history map is $\mathbf W_v\mathbf v_c(q)+\mathbf b_v$; the query projection is $\mathbf W_q\boldsymbol\eta_c(q)$, without a bias. Both are shared across channels. For local width $H$, $\mathbf W_v\in\mathbb R^{H\times(3K+7)}$, $\mathbf W_q\in\mathbb R^{H\times3}$, and $\mathbf b_v\in\mathbb R^H$. Lags $\boldsymbol\ell_c$ and flags $\mathbf u_c$ retain all $K$ slots; $D_c(q)$ is log-compressed elapsed time since $t_c^*$. The history input depends on $q$ through $\mathbf e_c(q)$. The projection sum $\mathbf r_c(q)$ enters CJD.

\subsection{Channel-conditioned joint decoding}
\label{sec:cjd}

\paragraph{Channel modulation and residual block.}
Channel modulation applies a scalar scale offset $\gamma_c$ and shift $\beta_c$ to the shared projection $\mathbf r_c(q)$. A SiLU activation and shared Linear--SiLU residual block produce the \emph{Local representation}:
\begin{equation}
\begin{aligned}
\mathbf{h}_c^{(0)}(q)&=\operatorname{SiLU}\!\left(
 (1+\gamma_c)\mathbf{r}_c(q)+\beta_c\mathbf{1}_H\right),\\
\mathbf{h}_c(q)&=\mathbf{h}_c^{(0)}(q)
 +\operatorname{SiLU}\!\left(\mathbf{W}_h\mathbf{h}_c^{(0)}(q)+\mathbf{b}_h\right).
\end{aligned}
\label{eq:local-modulation}
\end{equation}
Here $\mathbf{1}_H$ is the all-ones vector, $\operatorname{SiLU}(x)=x\sigma(x)$, $\mathbf W_h\in\mathbb R^{H\times H}$, and $\mathbf b_h\in\mathbb R^H$. The skip connection starts at the first SiLU output $\mathbf h_c^{(0)}(q)$ and adds the Linear--SiLU output. Only $\gamma_c$ and $\beta_c$ are channel-specific; the residual block is shared.

\paragraph{Context projection.}
The \emph{Joint descriptors} input concatenates six historical coordinates per variable. Context projection followed by SiLU gives the shared context:
\begin{equation}
\begin{aligned}
\mathbf{p}&=\mathop{\operatorname{concat}}_{c=1}^{C}
 [a_c,\mu_c,r_c,\ell_{c1},d_{c2},d_{c3}],\\
\mathbf{z}&=\operatorname{SiLU}(\mathbf{W}_p\mathbf{p}+\mathbf{b}_p).
\end{aligned}
\label{eq:context}
\end{equation}
The context width is $R$, with $\mathbf W_p\in\mathbb R^{R\times6C}$ and $\mathbf b_p\in\mathbb R^R$. We require $K\geq3$; unavailable deviations are zero. The same $\mathbf z$ serves every channel and query in a record.

\paragraph{Age-count gate.}
Elapsed time $D_c(q)$ and inverse count $r_c$ determine how much shared context reaches each query:
\begin{equation}
\begin{aligned}
\alpha_c(q)&=\sigma\!\left(\zeta_{c0}+\zeta_{c1}D_c(q)+\zeta_{c2}r_c\right),\\
\mathbf{z}_c(q)&=\alpha_c(q)\mathbf{z},
\end{aligned}
\label{eq:context-gate}
\end{equation}
The three $\zeta_{cr}$ coefficients are learned per channel. This sigmoid gate is a scalar, broadcast over all $R$ context coordinates at the multiplication node. Its output $\mathbf z_c(q)$ is the \emph{Gated context} group.

\paragraph{Joint features and channel readout.}
The \emph{Joint features} block concatenates the local representation, direct statistics, and gated context:
\begin{equation}
\mathbf{f}_c(q)=
[\mathbf{h}_c(q);\mathbf{m}_c^{(4)};\mathbf{z}_c(q)],
\label{eq:joint-features}
\end{equation}
The \emph{Direct statistics} group $\mathbf m_c^{(4)}$ selects the first four entries of $\mathbf m_c$, including $a_c$, along the direct path in Figure~\ref{fig:tress-framework}. The resulting $F=H+R+4$ coordinates enter one channel readout:
\begin{equation}
\widehat{y}_c(q)=(\mathbf{w}_c+\mathbf{e}_a)^\top\mathbf{f}_c(q)+b_c^{\mathrm{out}}.
\label{eq:forecast}
\end{equation}
Here $\mathbf w_c\in\mathbb R^F$ and $b_c^{\mathrm{out}}\in\mathbb R$ are learned, and the fixed one-hot vector $\mathbf e_a\in\mathbb R^F$ selects $a_c$. The latest-value coefficient is thus learned around one, with weight decay applied to $\mathbf w_c$. The three feature groups provide corrections to this persistence forecast.

\begin{algorithm}[t]
\caption{TRESS forecasting for one record}
\label{alg:tress}
\small
\begin{algorithmic}[1]
\REQUIRE History $(\mathbf X,\mathbf M,\mathbf t)$ and query times $\mathbf q$
\ENSURE Forecasts $\widehat{\mathbf Y}\in\mathbb R^{Q\times C}$
\STATE Find $t_0,t_{\min}$ and set $S=\max(t_0-t_{\min},0.05)$
\FOR{each channel $c$}
  \STATE Gather $K$ slots; set $a_c,t_c^*$ and $\mathbf u_c$
  \STATE Center events to $\mathbf d_c$; form $\boldsymbol\ell_c,\mathbf m_c$
\ENDFOR
\STATE Form $\mathbf p$; context projection and SiLU give $\mathbf z$ by \eqref{eq:context}
\FOR{each query $q$ and channel $c$}
  \STATE Event gate: form $\mathbf e_c(q)$ by \eqref{eq:event-gate}
  \STATE Sum history and query projections: $\mathbf r_c(q)$ by \eqref{eq:shared-projection}
  \STATE Channel modulation and residual block: $\mathbf h_c(q)$ by \eqref{eq:local-modulation}
  \STATE Age-count gate: form $\mathbf z_c(q)$ by \eqref{eq:context-gate}
  \STATE Concatenate $\mathbf h_c(q),\mathbf m_c^{(4)},\mathbf z_c(q)$ to $\mathbf f_c(q)$
  \STATE Channel readout: predict $\widehat y_c(q)$ by \eqref{eq:forecast}
\ENDFOR
\STATE \textbf{return} $\widehat{\mathbf Y}=[\widehat y_c(q)]_{q\in\mathbf q,\,c=1,\ldots,C}$
\end{algorithmic}
\end{algorithm}

\subsection{Training and model size}
\label{sec:training}

For a minibatch $\mathcal{B}$, let $Y_{bqc}$ and $N_{bqc}$ denote the target value and its binary observation mask for record $b$, query index $q$, and variable $c$. We minimize the observed-entry squared error,
\begin{equation}
\mathcal{L}=
\frac{\sum_{b\in\mathcal{B}}\sum_{q,c}N_{bqc}
  (\widehat{Y}_{bqc}-Y_{bqc})^2}
 {\max\!\left(\sum_{b\in\mathcal{B}}\sum_{q,c}N_{bqc},1\right)}.
\label{eq:training-loss}
\end{equation}
Event gathering and historical statistics depend only on the input record and fixed numerical hyperparameters and can be cached during training. Future values and their masks enter only the loss.

All four datasets use one residual block. For fixed $K$, $H$, and $R$, the parameter count grows linearly with the number of channels, independently of history length and query count. The exact parameter count is
\begin{equation}
\begin{aligned}
P={}&H^2+H(3K+12)+3(K-1)+R\\
   &+C(H+7R+10).
\end{aligned}
\label{eq:parameters}
\end{equation}

\section{Representing a history for forecast queries}
\label{sec:state}
Consider a table of measurements with columns for record, variable, timestamp, and value. A forecast request specifies a record, the end of its observation window, and one or more future times. A batch of requests may share the record and the observation window while differing in forecast time. Two properties of TRESS are useful for this workload. Its history inputs can be recovered from a bounded summary of the observed rows, and the same summary can serve every query on that snapshot. We describe the summary here to distinguish the cost of reading the data from the cost of evaluating the learned model.

\subsection{What must be kept from the history?}
For each variable $c$, retain its $K$ most recent observed timestamp--value pairs. Five aggregate fields describe the entire window. Writing $I_c$ for the observed row indices of variable $c$, these fields are
\begin{equation}
 \begin{aligned}
 n_c&=|I_c|,& s_c&=\sum_{i\in I_c}x_i,& w_c&=\sum_{i\in I_c}x_i^2,\\
 v_c&=\min_{i\in I_c}x_i,& b_c&=|\{i\in I_c:x_i=v_c\}|.&&
 \end{aligned}
\label{eq:state-agg}
\end{equation}
All five fields range over observed rows for this record and variable. At the record level keep the earliest and latest observed timestamps. The count distinguishes an empty history from a history of measured zeros. Empty groups follow the same padding convention as the batch encoder.

These fields recover every history-dependent input to TRESS in exact arithmetic for the same event ordering. The two sums give the mean and population variance, and $b_c/n_c$ gives the fraction of observations at the minimum whenever $n_c>0$. The retained pairs provide the latest value, the recent mean, the deviations, and their timestamps. Combining timestamps with the record endpoints gives the lags on the same scale as the batch encoder. In particular, an old observation that is absent from the recent buffer can change the mean, the minimum, or the historical span. Dropping that observation entirely would implement a different predictor.

For a fixed model, the state requires $O(CK+C)$ scalar fields per record, independent of the number of earlier measurements and the number of requested horizons. This bound concerns the input state for forecasting. It excludes the underlying event table, model parameters, predictions, and temporary tensors used during evaluation. The reported GPU benchmarks include the full scan performed by the batch implementation.

\subsection{Combining partial histories}
Two disjoint sets of observations for the same record can be summarized separately. Counts and sums add. The smaller of the two minima becomes the new minimum, and each partition contributes its count at the minimum only if its own minimum equals the new one. For example, if both partitions contain the value zero as their minimum, their counts of zeros add. Keeping only the count from the partition that supplied the most recent zero would change the feature $\pi_c$.

The recent-event buffer of the union is obtained by selecting the $K$ most recent pairs from the two buffers. Any discarded event already has at least $K$ later events in its own partition, so it cannot enter the joint buffer. Record endpoints combine by minimum and maximum. These operations also allow an insertion to update a precomputed summary. A row that arrives late may change any of the historical aggregates, even if its time falls outside the retained buffer. For repeated times within a variable, a persistent event identifier supplies a common tie order at construction and merge time. The benchmark tensor representation has one value per variable at a given row. Its top-$K$ operator does not specify how different values with identical timestamps should be ordered.

The scope of reuse is a fixed observation window. For an earlier cutoff, the more recent summary can contain observations that the query must not see. For a sliding window, expiring the only occurrence of the minimum can require the next smallest value, which this insertion state does not store. A service can build a new summary from the selected window or retain extra indexed state for expiry. The accuracy experiments use the supplied windows and masks.

\paragraph{Independent check against grouped queries.}
We implemented the merge rules and compared their output to SQL aggregates and an ordered top-$K$ query over the original event rows in SQLite. Each run generated histories for 19 records and 12 variables. Group sizes included zero, one, $K-1$, $K$, $K+1$, and 96 observations. Values were multiples of one quarter so sums and sums of squares were exact under the precision used. The input included repeated minima, measured zeros, and timestamps shared by distinct events. We randomly assigned rows to 1, 4, or 16 partitions and merged them in random order. With $K\in\{8,9,16\}$ and 12 random seeds, all 24,624 group comparisons agreed on counts, moments, minimum and its multiplicity, and the ordered recent events. This is a correctness test of the summary under its stated tie rule. The hardware timings use the native tensor implementation.

\section{Experiments}
\label{sec:experiments}

\begin{table*}[t]
\centering
\caption{Forecasting performance on four IMTS datasets (mean $\pm$ standard deviation). Bold and underline mark the lowest and second-lowest reported means. $\dagger$ denotes PhysioNet with a repaired test set, including reproduced baselines. Other baseline entries follow the reference comparison. TRESS results are averaged over five training seeds.}
\label{tab:accuracy}
\begingroup
\small
\setlength{\tabcolsep}{2pt}
\renewcommand{\arraystretch}{1.15}
\resizebox{\textwidth}{!}{%
\begin{tabular}{@{}l!{\color{black!40}\vrule width .35pt}cc!{\color{black!40}\vrule width .35pt}cc!{\color{black!40}\vrule width .35pt}cc!{\color{black!40}\vrule width .35pt}cc@{}}
\toprule
\bfseries Dataset & \multicolumn{2}{c}{\bfseries HumanActivity} & \multicolumn{2}{c}{\bfseries USHCN} & \multicolumn{2}{c}{\bfseries PhysioNet $^\dagger$} & \multicolumn{2}{c}{\bfseries MIMIC} \\
\midrule
\bfseries Metric & \bfseries MSE $\downarrow$ & \bfseries MAE $\downarrow$ & \bfseries MSE $\downarrow$ & \bfseries MAE $\downarrow$ & \bfseries MSE $\downarrow$ & \bfseries MAE $\downarrow$ & \bfseries MSE $\downarrow$ & \bfseries MAE $\downarrow$ \\
\midrule
PrimeNet
& 4.2507$\pm$0.0041
& 1.7018$\pm$0.0011
& 0.4930$\pm$0.0015
& 0.4954$\pm$0.0018
& 0.8128$\pm$0.00004
& 0.6977$\pm$0.00006
& 0.9073$\pm$0.0001
& 0.6614$\pm$0.0001 \\

NeuralFlows
& 0.1722$\pm$0.0090
& 0.3150$\pm$0.0094
& 0.2087$\pm$0.0258
& 0.3157$\pm$0.0187
& 0.3910$\pm$0.00303
& 0.4360$\pm$0.00201
& 0.6085$\pm$0.0101
& 0.5306$\pm$0.0066 \\

CRU
& 0.1387$\pm$0.0073
& 0.2607$\pm$0.0092
& 0.2168$\pm$0.0162
& 0.3180$\pm$0.0248
& 0.5917$\pm$0.00340
& 0.5604$\pm$0.00199
& 0.5895$\pm$0.0092
& 0.5151$\pm$0.0048 \\

mTAN
& 0.0993$\pm$0.0026
& 0.2219$\pm$0.0047
& 0.5561$\pm$0.2020
& 0.5015$\pm$0.0968
& 0.3711$\pm$0.00459
& 0.4231$\pm$0.00356
& 0.9408$\pm$0.1126
& 0.6755$\pm$0.0459 \\

SeFT
& 1.3786$\pm$0.0024
& 0.9762$\pm$0.0007
& 0.3345$\pm$0.0022
& 0.4083$\pm$0.0084
& 0.7912$\pm$0.00223
& 0.6860$\pm$0.00079
& 0.9230$\pm$0.0015
& 0.6628$\pm$0.0008 \\

GNeuralFlow
& 0.3936$\pm$0.1585
& 0.4541$\pm$0.0841
& 0.2205$\pm$0.0421
& 0.3286$\pm$0.0412
& 0.8433$\pm$0.15458
& 0.6791$\pm$0.06523
& 0.8957$\pm$0.0209
& 0.6450$\pm$0.0072 \\

GRU-D
& 0.1893$\pm$0.0627
& 0.3253$\pm$0.0485
& 0.2097$\pm$0.0493
& 0.3045$\pm$0.0305
& 0.3270$\pm$0.00157
& 0.3917$\pm$0.00091
& 0.4759$\pm$0.0100
& 0.4526$\pm$0.0055 \\

Raindrop
& 0.0916$\pm$0.0072
& 0.2114$\pm$0.0072
& 0.2035$\pm$0.0336
& 0.3029$\pm$0.0264
& 0.3330$\pm$0.00157
& 0.3965$\pm$0.00138
& 0.6754$\pm$0.1829
& 0.5444$\pm$0.0868 \\

Warpformer
& 0.0449$\pm$0.0010
& 0.1228$\pm$0.0018
& 0.1888$\pm$0.0598
& 0.2939$\pm$0.0591
& 0.2888$\pm$0.00141
& 0.3556$\pm$0.00095
& 0.4302$\pm$0.0035
& 0.4025$\pm$0.0014 \\

tPatchGNN
& 0.0443$\pm$0.0009
& 0.1247$\pm$0.0031
& 0.1885$\pm$0.0403
& 0.3084$\pm$0.0479
& 0.2923$\pm$0.00291
& 0.3595$\pm$0.00321
& 0.4431$\pm$0.0115
& 0.4077$\pm$0.0088 \\

GraFITi
& 0.0437$\pm$0.0005
& 0.1221$\pm$0.0017
& 0.1691$\pm$0.0093
& 0.2777$\pm$0.0248
& \underline{0.2873}$\pm$0.00213
& \underline{0.3531}$\pm$0.00214
& 0.4359$\pm$0.0455
& 0.4142$\pm$0.0297 \\

APN
& \underline{0.0421}$\pm$0.0001
& \underline{0.1159}$\pm$0.0006
& \underline{0.1590}$\pm$0.0137
& \underline{0.2611}$\pm$0.0167
& 0.2947$\pm$0.00310
& 0.3610$\pm$0.00610
& \underline{0.4292}$\pm$0.0027
& \underline{0.4016}$\pm$0.0016 \\

\midrule

\textbf{TRESS}
& \textbf{0.0405}$\pm$0.0001
& \textbf{0.1142}$\pm$0.0003
& \textbf{0.1513}$\pm$0.0002
& \textbf{0.2597}$\pm$0.0003
& \textbf{0.2844}$\pm$0.0002
& \textbf{0.3502}$\pm$0.0002
& \textbf{0.3862}$\pm$0.0008
& \textbf{0.3740}$\pm$0.0004 \\
\bottomrule
\end{tabular}%
}
\endgroup
\end{table*}

\subsection{Experimental setup}
\label{sec:setup}
We use the APN training, validation, and test partitions for HumanActivity, USHCN, PhysioNet, and MIMIC. PhysioNet retains the original training and validation caches and uses the repaired test cache. The architecture is shared across all four datasets. For each of five seeds (2024--2028), we evaluate the checkpoint with the lowest validation MSE. Test errors are pooled over observed targets within a seed, then reported as the mean and sample standard deviation across seeds. PhysioNet baseline results are reproduced on the \textbf{repaired dataset \cite{ma2026ctlite}}; other baseline values follow the reference comparison, with APN under its own settings.

\paragraph{Data and configurations.}
Table~\ref{tab:dataset-config} gives the fixed data partitions and model settings. We train with Adam, with batches of 128 records, gradient norms clipped at one, and a maximum of 160 epochs. Training stops after 25 epochs without improvement in validation MSE.
\begin{table*}[t]
\caption{Data partitions and TRESS configurations. $C$ counts variables, $K$ is the event budget, $H$ and $R$ are the widths of the local and context representations, and $\kappa$ controls the initial event gate.}
\label{tab:dataset-config}
\centering
\begin{tabular}{lcccccccccc}
\toprule
Dataset & Train & Validation & Test & $C$ & $K$ & $H$ & $R$ & $\kappa$ & Learning rate & Weight decay \\
\midrule
HumanActivity & 949 & 193 & 218 & 12 & 8 & 12 & 12 & 0.50 & 0.003 & 0.001 \\
USHCN & 901 & 101 & 112 & 5 & 9 & 16 & 15 & 0.50 & 0.010 & 0.075 \\
PhysioNet & 9,703 & 1,079 & 1,199 & 36 & 16 & 20 & 12 & 1.00 & 0.001 & 0.001 \\
MIMIC & 17,212 & 1,913 & 2,125 & 96 & 9 & 16 & 9 & 0.25 & 0.001 & 0.001 \\
\bottomrule
\end{tabular}
\end{table*}

Our comparison includes methods developed for irregular observations, including PrimeNet~\cite{primenet}, SeFT~\cite{horn2020seft}, GNeuralFlow~\cite{gneuralflow}, Raindrop~\cite{raindrop}, and Warpformer~\cite{warpformer}. We compare against persistence on exactly the same targets in Section~\ref{sec:age-diagnostic}.

\subsection{Forecasting performance}
\label{sec:accuracy}
TRESS ranks first among the methods in Table~\ref{tab:accuracy} for both mean test MSE and mean test MAE on all four datasets. On MIMIC, MSE falls from 0.4292 to 0.3862 and MAE from 0.4016 to 0.3740, reductions of 10.02\% and 6.87\% relative to APN. HumanActivity improves by 3.89\% in MSE and 1.43\% in MAE. On PhysioNet, the reductions relative to GraFITi are 1.00\% and 0.81\%. On USHCN, TRESS obtains MSE 0.1513 and MAE 0.2597, below APN's reference means by 4.85\% and 0.55\%.

\begin{table*}[t]
\centering
\caption{Component ablations on four IMTS datasets (test mean $\pm$ sample standard deviation over five seeds). Each variant removes one of the three readout feature groups and uses TRESS's selected hyperparameters. Removing direct statistics also removes the latest-value coordinate and its unit coefficient prior. Values are rounded to four decimal places. $\dagger$ marks the repaired PhysioNet test set.}
\label{tab:ablation}
\begingroup
\small
\setlength{\tabcolsep}{2pt}
\renewcommand{\arraystretch}{1.15}
\resizebox{\textwidth}{!}{%
\begin{tabular}{@{}l!{\color{black!40}\vrule width .35pt}cc!{\color{black!40}\vrule width .35pt}cc!{\color{black!40}\vrule width .35pt}cc!{\color{black!40}\vrule width .35pt}cc@{}}
\toprule
\bfseries Dataset & \multicolumn{2}{c}{\bfseries HumanActivity} & \multicolumn{2}{c}{\bfseries USHCN} & \multicolumn{2}{c}{\bfseries PhysioNet $^\dagger$} & \multicolumn{2}{c}{\bfseries MIMIC} \\
\midrule
\bfseries Metric & \bfseries MSE $\downarrow$ & \bfseries MAE $\downarrow$ & \bfseries MSE $\downarrow$ & \bfseries MAE $\downarrow$ & \bfseries MSE $\downarrow$ & \bfseries MAE $\downarrow$ & \bfseries MSE $\downarrow$ & \bfseries MAE $\downarrow$ \\
\midrule
w/o local representation
& 0.0427$\pm$0.0000
& 0.1172$\pm$0.0003
& 0.1495$\pm$0.0000
& 0.2595$\pm$0.0001
& 0.2902$\pm$0.0003
& 0.3531$\pm$0.0002
& 0.3985$\pm$0.0007
& 0.3826$\pm$0.0013 \\

w/o direct statistics
& 0.0408$\pm$0.0004
& 0.1166$\pm$0.0010
& 0.1889$\pm$0.0018
& 0.2978$\pm$0.0048
& 0.2836$\pm$0.0004
& 0.3524$\pm$0.0020
& 0.3892$\pm$0.0014
& 0.3778$\pm$0.0010 \\

w/o gated context
& 0.0408$\pm$0.0000
& 0.1132$\pm$0.0001
& 0.1513$\pm$0.0001
& 0.2596$\pm$0.0003
& 0.2873$\pm$0.0002
& 0.3512$\pm$0.0003
& 0.4048$\pm$0.0004
& 0.3806$\pm$0.0003 \\

\midrule

TRESS
& 0.0405$\pm$0.0001
& 0.1142$\pm$0.0003
& 0.1513$\pm$0.0002
& 0.2597$\pm$0.0003
& 0.2844$\pm$0.0002
& 0.3502$\pm$0.0002
& 0.3862$\pm$0.0008
& 0.3740$\pm$0.0004 \\
\bottomrule
\end{tabular}%
}
\endgroup
\end{table*}

\subsection{Component ablations}
\label{sec:ablation}
We remove each of the three \emph{Joint features} groups in Figure~\ref{fig:tress-framework}: local representation $\mathbf h_c(q)$, direct statistics $\mathbf m_c^{(4)}$, and gated context $\mathbf z_c(q)$ (Table~\ref{tab:ablation}). Each variant deletes that group's readout coefficients and any layers used exclusively by it, then retrains the remaining model under the same settings. Retained parameters use paired initialization within each seed, and paired runs share the minibatch order.

\paragraph{Local representation.}
Removing the local representation deletes the QEL projections and event gate, together with CJD's channel modulation and residual block. MSE rises by 5.61\%, 2.02\%, and 3.18\% on HumanActivity, PhysioNet, and MIMIC, respectively; MAE also rises on all three. On USHCN, removal lowers MSE from 0.1513 to 0.1495 and MAE from 0.2597 to 0.2595.

\paragraph{Direct statistics.}
Direct statistics expose the current level and history summaries to the readout. Removing them raises MAE on every dataset and raises MSE on three. USHCN shows the largest degradation, with increases of 24.88\% in MSE and 14.69\% in MAE. PhysioNet has a small MSE reduction of 0.29\%, alongside a 0.63\% MAE increase. This intervention removes both the four statistics and the unit prior on the latest-value coordinate, so it measures their joint contribution.

\paragraph{Gated context.}
Removing the context projection and age-count gate raises MIMIC MSE by 4.81\% and MAE by 1.76\%, the largest changes among the four datasets. Both errors also rise on PhysioNet. HumanActivity shows higher MSE and lower MAE, while both USHCN errors decrease by less than 0.03\%. Cross-variable context therefore contributes most strongly on MIMIC, with smaller or mixed effects elsewhere.

\subsection{Forecasts stratified by observation age}
\label{sec:age-diagnostic}
How much does TRESS gain over carrying a variable's last observation forward? We replayed the five released checkpoints per dataset on the original test records. The persistence forecast repeats the latest observed value of the target variable at every query. For a variable absent from the history, it predicts zero in the benchmark's preprocessed coordinates, the same default used for TRESS's anchor. Both methods are scored on the same observed target entries. Across all 20 full-model checkpoints, the replayed MSE and MAE match the main experiment within $10^{-6}$.

\begin{table}[t]
\centering
\caption{Five-seed test errors against the persistence forecast. The target count is the number of observed scalar forecast values.}
\label{tab:persistence-age}
\begin{tabular}{@{}lcccrr@{}}
\toprule
& &\multicolumn{2}{c}{Persistence} & \multicolumn{2}{c}{TRESS}\\
\cmidrule(lr){3-4}\cmidrule(l){5-6}
Dataset & Targets & MSE & MAE & MSE & MAE\\
\midrule
HumanActivity & 5,910 & 0.0605 & 0.1225 & 0.0405 & 0.1142\\
USHCN & 368 & 0.3640 & 0.3291 & 0.1513 & 0.2597\\
PhysioNet & 27,008 & 0.3985 & 0.3919 & 0.2844 & 0.3502\\
MIMIC & 23,015 & 0.5702 & 0.4254 & 0.3862 & 0.3740\\
\bottomrule
\end{tabular}
\end{table}

Table~\ref{tab:persistence-age} shows that the learned correction lowers both errors on every dataset. The reduction in MSE ranges from 28.6\% on PhysioNet to 58.4\% on USHCN. We next ask when the shared context contributes to this correction. We compare TRESS with the independently retrained variant without gated context, pairing their predictions by target and initialization. Both evaluation paths use the full test set.

\begin{figure}[t]
\centering
\includegraphics[width=\columnwidth]{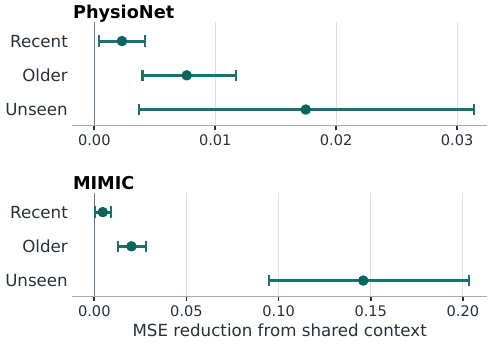}
\caption{Test MSE of the retrained model without context minus that of TRESS. Positive values favor the full model. A recent variable was last observed less than one quarter of the history span before its endpoint, an older variable was last observed at least one quarter-span before the endpoint, and an unseen variable has no history in that record. Bars are 95\% paired record-bootstrap intervals. Panels use separate horizontal scales.}
\Description{Paired changes in MSE by age of the last observation for two clinical cohorts.}
\label{fig:context-age}
\end{figure}

For every observed test target we assign its variable to one of three groups using the history available at prediction time. With normalized age $(t_0-t_c^*)/S$, the recent and older groups use $[0,0.25)$ and $[0.25,1]$, respectively. The same quarter-span threshold is used in the training-history analysis. Targets whose variable has never been observed form a third group. For each group, we average the seed-specific MSE differences; 2,000 paired bootstrap samples of whole test records give percentile intervals conditional on the fitted checkpoints. HumanActivity has 5,910 recent targets and none in the other groups. USHCN has only three older targets, all from one record, and no unseen targets. The clinical datasets support comparisons across all three groups.

In PhysioNet the three groups contain 24,434, 2,289, and 285 targets, respectively. The MSE reduction with the context branch grows from 0.0023 for recent targets to 0.0076 for older targets and 0.0175 for unseen targets (Figure~\ref{fig:context-age}). MIMIC exhibits a larger difference. The corresponding reductions are 0.0046, 0.0202, and 0.1460 across 16,252, 5,039, and 1,724 targets. For the unseen group, MIMIC MSE falls from 1.0962 to 0.9502 when shared context is available. These groups differ in variable and patient composition. The ordering therefore describes which groups benefited within these test cohorts.

\begin{figure*}[t]
\centering
\includegraphics[width=\textwidth]{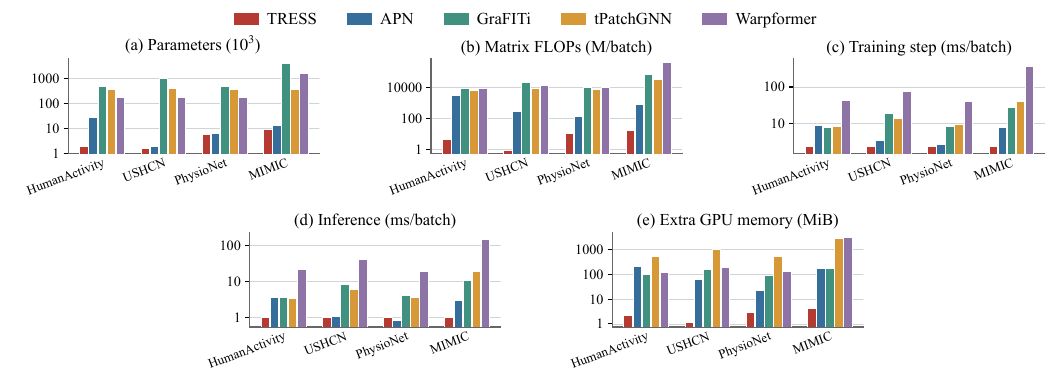}
\caption{Measured computational costs on an RTX 3090 at batch size 32; lower is better. Each panel uses a logarithmic vertical axis. Training-step time includes forward, masked MSE, backward, and Adam; inference starts from raw histories and includes model-specific input conversion. Matrix FLOPs count profiled matrix products and convolutions rather than all operations. Memory is the extra eager peak beyond resident model and input tensors. Section~\ref{sec:efficiency} gives the measurement protocol.}
\Description{Five panels compare parameters, matrix operations, training time, inference time and extra peak memory by dataset and model.}
\label{fig:efficiency}
\end{figure*}

\subsection{How much does the history before the latest value contribute?}
\label{sec:history-thinning}
We thin the test histories while preserving the most recent observation of every variable that has been measured. This intervention holds each variable's latest value and timestamp fixed, so the persistence baseline makes exactly the same prediction at every retention rate. Each earlier observed entry receives a uniform random number. We retain entries with numbers below $p$, using the same numbers at $p=1/2$ and $p=1/4$ so that the masks are nested. We also evaluate the complete history ($p=1$) and the latest observation alone for each variable ($p=0$). These sparsified records have the same queries and observed targets as the originals. TRESS recomputes all features from the retained history, including counts, summary statistics, and the normalizing span $S$. The common historical endpoint is preserved.

\begin{table}[t]
\centering\small
\caption{Test MSE after independently thinning earlier observations. Each variable retains its latest measurement. The persistence column applies at every retention probability $p$.}
\label{tab:history-thinning}
\begin{tabular}{@{}lccccc@{}}
\toprule
& \multicolumn{4}{c}{TRESS, earlier-observation retention $p$} & \\
\cmidrule(lr){2-5}
Dataset & $1$ & $1/2$ & $1/4$ & $0$ & Persistence \\
\midrule
HumanActivity & 0.0405 & 0.0430 & 0.0454 & 0.0576 & 0.0605\\
USHCN         & 0.1513 & 0.1541 & 0.1680 & 0.2812 & 0.3640\\
PhysioNet     & 0.2844 & 0.2913 & 0.3010 & 0.3450 & 0.3985\\
MIMIC         & 0.3862 & 0.3955 & 0.4064 & 0.4131 & 0.5702\\
\bottomrule
\end{tabular}
\end{table}

Table~\ref{tab:history-thinning} averages observed-target MSE over the five frozen models and, at the two intermediate rates, three independently drawn masks. Training used the original sampling patterns. Mean MSE rises as the input loses earlier observations on all four datasets. The averages on USHCN are particularly sensitive to the mask: at $p=1/4$, individual mask means across the five models range from 0.1416 to 0.2022. With only the latest measurement of each variable, TRESS still has lower MSE than persistence on every dataset. The largest remaining gap is on MIMIC, where the two models obtain 0.4131 and 0.5702, respectively. Here, TRESS can use the latest levels of other variables as well as the elapsed time to a query. 

The latest-only condition removes different fractions of the input in the four cohorts. In MIMIC, it preserves 79,942 of the 250,445 original test-history observations (31.9\%). The corresponding fractions are 9.4\% on PhysioNet, 4.4\% on HumanActivity, and 1.9\% on USHCN. Retention probabilities apply to observations preceding the anchor, so comparisons at a given $p$ control the sampling rule within a dataset. They do not equalize the amount of information available across datasets. None of the four test caches had a tie for the latest observed timestamp within a record and variable.

\begin{figure*}[t]
\centering
\includegraphics[width=\textwidth]{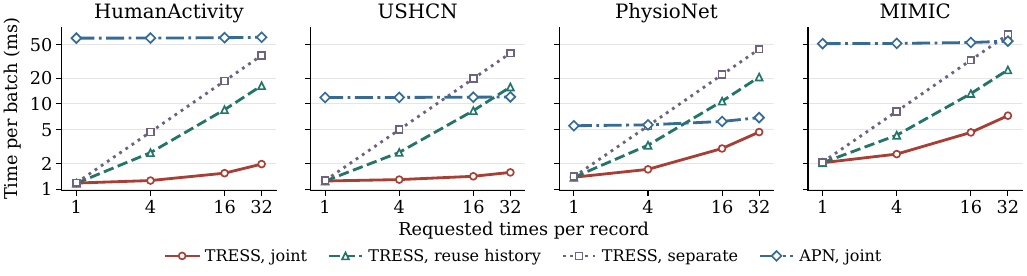}
\caption{Time to return forecasts for eight records as the number of requested times per record increases. Joint execution batches the times. Reuse history and separate execution issue them one at a time. Points are medians over three batches on a single CPU thread. Both axes use logarithmic scales.}
\Description{Latency for separate calls, cached histories, joint execution, and APN at increasing query counts in four datasets.}
\label{fig:query-scaling}
\end{figure*}

\subsection{What do observations outside the event buffer contribute?}
\label{sec:summary-window}
Thinning changes both the recent events and the statistics passed to the decoder. To isolate the contribution of measurements older than the event buffer, we replayed the five fitted models on the original test records. For each variable we kept its latest $K$ events, their timestamps, and the original time span used to normalize lags. We then recomputed the mean, minimum, standard deviation, and fraction of values at the minimum using only those events, wherever the model uses these quantities. A second pass also replaced the full observation count $n_c$ with $\min(n_c,K)$. All three passes used the same weights, forecast times, and target mask.

\begin{table}[t]
\centering
\caption{Test MSE / MAE when the window used for statistics changes. Full uses the complete history. Recent uses the latest $K$ values with the original count. Capped also limits the count to $K$. Entries average the same five training seeds.}
\label{tab:summary-window}
\setlength{\tabcolsep}{3pt}
\begin{tabular}{@{}lccc@{}}
\toprule
Dataset & Full & Recent & Capped \\
\midrule
HumanActivity & 0.0405 / 0.1142 & 0.0404 / 0.1134 & 0.0404 / 0.1134 \\
USHCN         & 0.1513 / 0.2597 & 0.2289 / 0.3096 & 0.2288 / 0.3096 \\
PhysioNet     & 0.2844 / 0.3502 & 0.2845 / 0.3497 & 0.2845 / 0.3498 \\
MIMIC         & 0.3862 / 0.3740 & 0.3860 / 0.3735 & 0.3860 / 0.3737 \\
\bottomrule
\end{tabular}
\end{table}

The strongest effect appears on USHCN, where the nine-event buffers cover only 17.0\% of historical measurements. Restricting the value statistics raises MSE by 0.0776 and MAE by 0.0499 (Table~\ref{tab:summary-window}). The MSE increase has a 95\% paired bootstrap interval of $[0.0380,0.1197]$ across 2,000 resamples of whole test records, with the five fits held fixed. Capping the count changes USHCN MSE by less than 0.0001. The older measurements thus inform forecasts through the values summarized beyond the buffer, even when their individual timestamps are no longer retained.

On HumanActivity, PhysioNet, and MIMIC, restricting the value statistics changes mean MSE by at most 0.0003, and MAE falls slightly on all three. Earlier observations still matter under thinning (Table~\ref{tab:history-thinning}), which can also remove events that would have occupied a recent slot. Preserving those slots yields a different picture of how the fitted models use their histories.

\subsection{Parameter count and computational cost}
\label{sec:efficiency}
We benchmark all five methods on the same cached records, timestamps, masks, and forecast queries at batch size 32. Measurements include each model's input conversion and use ordinary PyTorch execution on one RTX 3090. Training steps use training records; inference uses test records. GraFITi and tPatchGNN use their official implementations with APN-reference configurations. Warpformer uses APN's forecasting adaptation. Figure~\ref{fig:efficiency} shows the costs.

TRESS uses 1,574--9,457 parameters, the fewest among the five methods on each dataset. Its extra inference memory is 1.20--4.45 MiB; APN requires 7.6--98.9 times as much. TRESS also has the lowest matrix-operation count and measured training-step time on all four datasets. Inference is 3.60, 1.07, and 2.93 times faster than APN on HumanActivity, USHCN, and MIMIC. On PhysioNet, TRESS takes 1.015 ms versus APN's 0.840 ms despite requiring fewer matrix FLOPs.

\paragraph{Measurement protocol.}
We benchmark the beginning, middle, and end batches of each split. Each measurement uses five warm-up calls and five groups of ten timed calls, synchronized before and after each group. Reported per-call times are medians within a batch and then across the three batches. We use PyTorch 2.5.1, CUDA 12.4, four CPU threads, and FP32 with TF32 disabled. A training step comprises the forward pass from raw input, masked loss, backpropagation, and an Adam update. A forward pass includes model-specific input conversion. Extra inference memory measures the largest increase in allocated GPU memory across the three batches. The profiler counts matrix products and convolutions with two operations per multiply--add.

\begin{figure*}[!t]
\centering
\includegraphics[width=\textwidth]{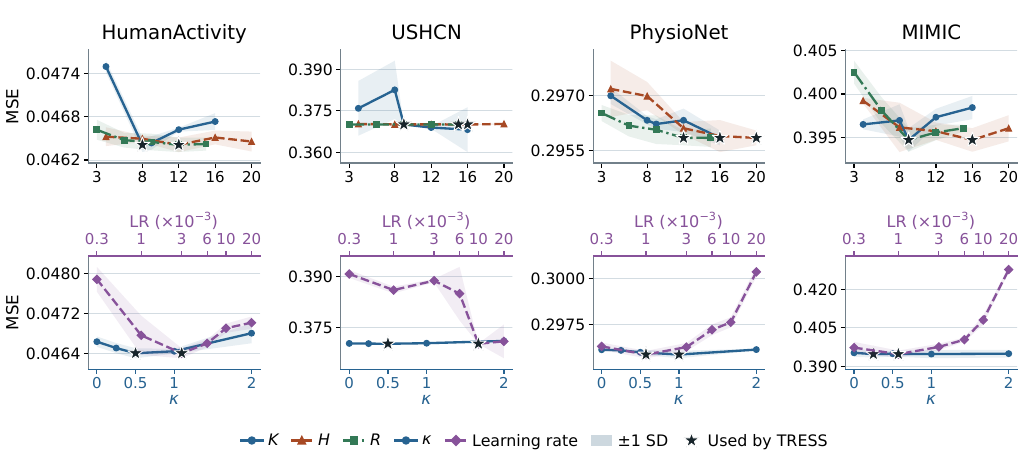}
\caption{Validation MSE as one parameter varies around the selected TRESS configuration. Columns correspond to datasets. Lines show three-seed means, bands span one sample standard deviation, and stars mark the settings used by TRESS. The upper row overlays $K$, $H$, and $R$ on a common numerical axis. The lower row overlays $\kappa$ (bottom axis, linear) and learning rate (top axis, logarithmic, in units of $10^{-3}$), sharing one MSE axis. }
\Description{Validation MSE curves for five hyperparameters on four datasets. Stars mark the settings used to evaluate TRESS.}
\label{fig:sensitivity-merged}
\end{figure*}

\subsection{Serving multiple forecast times}
\label{sec:query-scaling}
A forecast request can specify one future time or an entire trajectory. If several requests concern the same record, TRESS can reuse the output of its history encoder until a new observation arrives. We measured three ways to serve the same $Q$ times. \emph{Separate} runs the complete model for each time. \emph{Reuse history} runs the encoder once and invokes the decoder for each time, including the first encoding in the measured cost. \emph{Joint} passes all times to the native model in one call. We also measured APN with its native joint interface. These comparisons separate the benefit of reusing the history from the additional benefit of evaluating the decoder in a batch.

We selected the first eight records from each of the three archived test batches per dataset. For each record we placed $Q\in\{1,4,16,32\}$ equally spaced queries between the history endpoint (excluded) and a quarter of the history span $S$ beyond that endpoint (included). These queries define a computational workload and require no target values. We used trained checkpoints in FP32 with one PyTorch thread pinned to a logical CPU on an Intel Core i7-12700KF. Timings start with the records and query tensors in memory and include all encoding, decoding, and output assembly. We randomized execution order within each block, warmed up for two blocks, then took the median of seven block means, each over three repetitions. Scheduling changed predictions by at most $1.5\times10^{-6}$ in absolute value across both models and all batches.

Figure~\ref{fig:query-scaling} shows that the two forms of reuse have distinct effects. At $Q=32$, encoding once reduced TRESS's time for separate requests by a factor of 2.12--2.60 across the four datasets. The cached tensors occupied 0.83--15.76\,KiB per record, excluding allocator and container overhead. When all 32 times were available together, batching the decoder was a further 3.44--10.05 times faster. On MIMIC, the corresponding batch times were 65.05\,ms for separate calls, 25.04\,ms with history reuse, and 7.27\,ms for joint execution. Both models amortized history processing across query times. On PhysioNet, for example, APN's joint time increased from 5.54 to 6.90\,ms between one and 32 times, while TRESS's increased from 1.38 to 4.67\,ms. The lower fixed cost of TRESS at one query thus coincided with a larger increase in latency over this range.

\subsection{Cost of padded history storage}
\label{sec:padded-history}
Recent events and full-history statistics occupy a fixed number of coordinates per variable. Constructing those coordinates still requires access to the input record. In a dense batch, reductions over historical values and the search for the most recent events both encounter rows allocated to padding. We isolate this cost on PhysioNet and MIMIC, the two clinical datasets with 36 and 96 variables, respectively.

We used the trained models, records, processor, and batch size from Section~\ref{sec:query-scaling}. We held four queries per record fixed and extended the history tensors to one, two, four, and eight times their original allocated length. Each appended row contained zeros for all values, timestamps, and observation indicators. Each observed event retained its value, time, and position. The padded tensors were already resident when timing began. We measured end-to-end inference for both models, TRESS's history encoder alone, and its decoder supplied with the corresponding cached history. We shuffled all 16 combinations of padding level and operation within each timing block and used the repetitions and aggregation from the query-count experiment.

\begin{table}[t]
\centering
\caption{Median inference time in milliseconds per batch with a fixed set of observations and four query times. $L$ counts all allocated history rows. Encode and decode are separate measurements of TRESS's components.}
\label{tab:padded-history}
\setlength{\tabcolsep}{7pt}
\begin{tabular}{lccccc}
\toprule
 & & \multicolumn{3}{c}{TRESS} & APN \\
\cmidrule(lr){3-5}
Dataset & $L$ & Full & Encode & Decode & Full \\
\midrule
PhysioNet & 36  & 1.73 & 0.71 & 0.98 & 5.68 \\
          & 72  & 1.84 & 0.84 & 0.97 & 8.54 \\
          & 144 & 2.06 & 1.06 & 0.97 & 16.75 \\
          & 288 & 2.45 & 1.44 & 0.98 & 31.49 \\
\midrule
MIMIC     & 72  & 2.62 & 1.27 & 1.29 & 43.16 \\
          & 144 & 3.20 & 1.89 & 1.29 & 93.97 \\
          & 288 & 4.47 & 3.08 & 1.28 & 218.73 \\
          & 576 & 6.44 & 5.11 & 1.28 & 430.45 \\
\bottomrule
\end{tabular}
\end{table}

Table~\ref{tab:padded-history} distinguishes input traversal from forecasting with an encoded record. With eight times as many allocated rows, TRESS's encoding latency increased by factors of 2.02 on PhysioNet and 4.03 on MIMIC. Cached decoding took 0.97--0.98\,ms on PhysioNet and 1.28--1.29\,ms on MIMIC. Every encoded tensor matched its counterpart at the original length exactly. The payload remained 8.87\,KiB per PhysioNet record and 15.76\,KiB per MIMIC record. Across both models, the largest absolute prediction change from padding was $4.77\times10^{-7}$.

For repeated requests over the same record, caching removes a scan whose cost depends on the physical batch layout. For a first request, grouping records of similar length would reduce the number of empty rows the encoder must examine. The measured changes arise solely from allocating additional rows at fixed information content. Adding observed events can also change the retained history and the forecast.

\subsection{Parameter sensitivity}
\label{sec:sensitivity}
We vary the event budget $K$, local width $H$, context width $R$, event-gate offset $\kappa$, and learning rate one at a time around the final configurations. Figure~\ref{fig:sensitivity-merged} shows validation MSE over seeds 2024--2026 for all 104 dataset--parameter settings. Other settings remain fixed within each sweep. Each plotted mean and standard deviation spans three seeds. Every sweep uses the same minimum-validation-MSE rule for choosing checkpoints.

The selected settings minimize mean validation MSE in nineteen of the twenty parameter--dataset sweeps. For USHCN's event count, increasing $K$ beyond 9 lowers mean validation MSE but increases variation across seeds: the standard deviation is $4.63\times10^{-5}$ at $K=9$ and $8.05\times10^{-3}$ at $K=16$. The USHCN context-width curve is nearly flat, spanning 0.0027\% of the selected MSE. MIMIC responds more strongly to context width, whose range is 1.99\%, compared with 0.95\% for event count. PhysioNet varies by less than 0.46\% in each of the four model-parameter sweeps. The selected learning rates minimize mean validation MSE on all four datasets. Increasing MIMIC's learning rate from 0.001 to 0.02 raises MSE from 0.3947 to 0.4276.

\section{Discussion}
\label{sec:discussion}
\paragraph{Which targets account for the gains?}
The distribution of test targets matters when interpreting an overall MSE. For a group $g$ containing $N_g$ observed targets, let $\Delta_g$ be the reduction in MSE from adding shared context to the retrained model. Then $N_g\Delta_g$ is the group's contribution to the total reduction in squared error. This decomposition uses the same target weights as the training objective. A rarely observed variable can contribute a large share of the difference between two models even when it accounts for few predictions.

In MIMIC, targets whose variable was absent from the history account for 7.5\% of observed test values and 58.8\% of the reduction in squared error due to the context branch. They occur in 473 test records. PhysioNet gives a different picture. Unseen variables have the largest per-target improvement there too, but they account for only 285 targets in 95 records. Their share of the reduction is 6.4\%. Recent variables account for 71.3\% of the reduction on PhysioNet. Thus, retaining cross-variable information helps an important group in both cohorts, while its effect on the reported overall metric depends strongly on how often that group occurs.

\begin{table}[t]
\caption{Clinical test MSE by the target variable's observation history. The middle predictor is TRESS retrained without gated context. Within each row, all three predictors are scored on the same targets.}
\label{tab:age-absolute}
\centering
\begin{tabular}{@{}llcccc@{}}
\toprule
Dataset & Group & Targets & Persist. & No context & TRESS\\
\midrule
PhysioNet & Recent & 24,434 & 0.3926 & 0.2799 & 0.2777\\
 & Older & 2,289 & 0.3959 & 0.2908 & 0.2832\\
 & Unseen & 285 & 0.9258 & 0.8936 & 0.8761\\
MIMIC & Recent & 16,252 & 0.5278 & 0.3557 & 0.3511\\
 & Older & 5,039 & 0.5018 & 0.3265 & 0.3063\\
 & Unseen & 1,724 & 1.1699 & 1.0962 & 0.9502\\
\bottomrule
\end{tabular}
\end{table}
Table~\ref{tab:age-absolute} gives the absolute errors behind Figure~\ref{fig:context-age}. On MIMIC, all three predictors have lower MSE on older targets than on recent ones. These groups contain different records and variables. The older group can therefore benefit more from context while having lower absolute error. Variables unseen in the history are the hardest to forecast for all three predictors on both clinical datasets.

\paragraph{What the error averages measure.}
The forecast target is a measured value at a requested time. Multiple measurements from one record can enter the average, while some records have no observed targets in the evaluation window. For example, 1,755 of the 2,125 MIMIC test records and 1,194 of the 1,199 PhysioNet test records contribute to the reported errors. Weighting all records equally would therefore answer a different question. We pool observed scalar targets when reporting MSE, and resample records rather than individual measurements when estimating uncertainty.

Using the paired record bootstrap from Section~\ref{sec:age-diagnostic}, the MSE advantage over persistence on MIMIC is 0.1840, with a 95\% interval of [0.1549, 0.2168]. The corresponding advantage is 0.1141 [0.0984, 0.1303] on PhysioNet and 0.0200 [0.0125, 0.0290] on HumanActivity. These intervals compare predictions for the same test records and checkpoints. In particular, the larger spread on MIMIC reflects variation across sampled records that the small standard deviations across training seeds in Table~\ref{tab:accuracy} do not capture. Both summaries are relevant when predicting for a new set of records.

\paragraph{Choosing an execution plan.}
Known forecast times can share one query tensor, batching decoder work across records and times. When the next request depends on the preceding answer, an encoded snapshot lets the client issue one query at a time. New observations invalidate the snapshot across channels because even a measurement of another variable can change the target's context and the common time origin.

\paragraph{Scope of the evaluation.}
We score the targets that were actually recorded. In a clinical record the decision to take a measurement can depend on a patient's condition, so errors on the observed targets do not estimate errors at all physically possible times. The age analysis preserves this sampling pattern. Random thinning answers a separate question about the same trained predictor under loss of earlier inputs. The paired intervals quantify variability across records given the fitted models. They do not establish that a particular delay in measurement causes cross-variable context to become useful. On USHCN, few targets have older observations, limiting what we can infer about forecasts from stale measurements.

\section{Conclusion}
\label{sec:conclusion}
We introduced TRESS, a compact forecasting model for multivariate time series in which variables are observed at different times. TRESS retains each variable's recent measurements at their original times and expresses them relative to its latest value. Query-conditioned event lifting weights these deviations by observation age and forecast horizon, allowing the same measurement to contribute differently to near-term and distant forecasts. Statistics of the full history carry information from earlier observations into this fixed-size representation. A channel-specific readout then combines local features, historical statistics, and context from the other variables to predict a correction to the latest value. Reusing the encoded history also lets successive forecast queries share the cost of reading the record. With fewer than 10K trainable parameters on each benchmark, TRESS recorded the lowest mean test MSE and MAE on all four datasets in our comparison.

\begin{acks}
This work was supported by the Deep Space Exploration Directed Cultivation Fund of USTC.
\end{acks}

\clearpage
\bibliographystyle{ACM-Reference-Format}
\bibliography{references}
\end{document}